%% file: main.tex
\documentclass[10pt,twocolumn,letterpaper]{article}

\usepackage[pagenumbers]{iccv} 


\usepackage{multirow}
\usepackage{makecell}
\usepackage{amssymb}
\usepackage[normalem]{ulem}
\usepackage{graphicx}
\usepackage{wrapfig} 
\usepackage{hyperref}
\useunder{\uline}{\ul}{}

\newcommand{\gray}[1]{\textcolor{gray}{\textit{#1}}}

\definecolor{iccvblue}{rgb}{0.21,0.49,0.74}
\hypersetup{
  pagebackref=true,
  breaklinks=true,
  colorlinks=true,
  allcolors=iccvblue
}

\def\paperID{7732} 
\def\confName{ICCV}
\def\confYear{2025}

\title{S$^3$POT: Contrast-Driven Face Occlusion Segmentation via Self-Supervised Prompt Learning}

\author{
Lingsong Wang$^{1,2}$ \quad
Mancheng Meng$^{2}$ \quad
Ziyan Wu$^{2}$ \quad
Terrence Chen$^{2}$ \quad
Fan Yang$^{2}$\thanks{ Corresponding author} \quad
Dinggang Shen$^{1, 2}$ \\
$^{1}$ShanghaiTech University \\
$^{2}$United Imaging Intelligence \\
}

\begin{document}
\maketitle
\input{sec/0_abstract}    
\input{sec/1_intro}

\input{sec/2_relatedwork}
\input{sec/3_method}
\input{sec/4_experiment}
\input{sec/5_conclusion}
{
    \small
    \bibliographystyle{ieeenat_fullname}
    \bibliography{main}
}

\end{document}

%% file: sec/0_abstract.tex
\begin{abstract}
Existing face parsing methods usually misclassify occlusions as facial components. This is because occlusion is a high-level concept, it does not refer to a concrete category of object. Thus, constructing a real-world face dataset covering all categories of occlusion object is almost impossible and accurate mask annotation is labor-intensive. To deal with the problems, we present S$^3$POT, a contrast-driven framework synergizing face generation with self-supervised spatial prompting, to achieve occlusion segmentation. The framework is inspired by the insights: 1) Modern face generators' ability to realistically reconstruct occluded regions, creating an image that preserve facial geometry while eliminating occlusion, and 2) Foundation segmentation models' (e.g., SAM) capacity to extract precise mask when provided with appropriate prompts. 
In particular, S$^3$POT consists of three modules: Reference Generation (RF), Feature enhancement (FE), and Prompt Selection (PS). First, a reference image is produced by RF using structural guidance from parsed mask. Second, FE performs contrast of tokens between raw and reference images to obtain an initial prompt, then modifies image features with the prompt by cross-attention. Third, based on the enhanced features, PS constructs a set of positive and negative prompts and screens them with a self-attention network for a mask decoder. The network is learned under the guidance of three novel and complementary objective functions without occlusion ground truth mask involved. Extensive experiments on a dedicatedly collected dataset demonstrate S$^3$POT's superior performance and the effectiveness of each module. Code is available at: \url{https://github.com/Bh-Johnny/S3SPOT}.

\end{abstract}

%% file: sec/1_intro.tex
\begin{figure}
    \centering
    \includegraphics[width=0.98\linewidth]{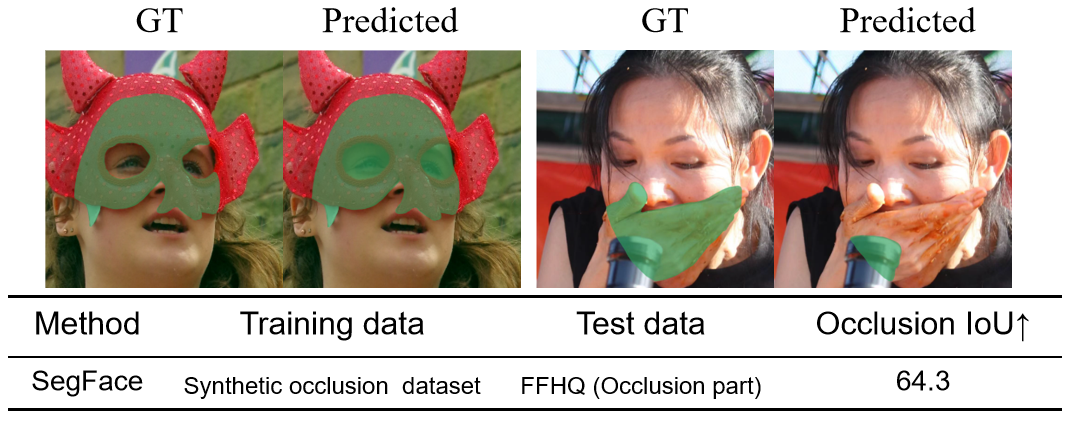}
    \caption{Qualitative and quantitative results of SegFace trained on a large-scale synthetic dataset~\cite{voo2022delving} and tested on our collected data.}
    \label{fig:front}
\end{figure}


\section{Introduction}
\label{sec:intro}
\begin{figure*}
    \centering
    \includegraphics[width=0.95\linewidth]{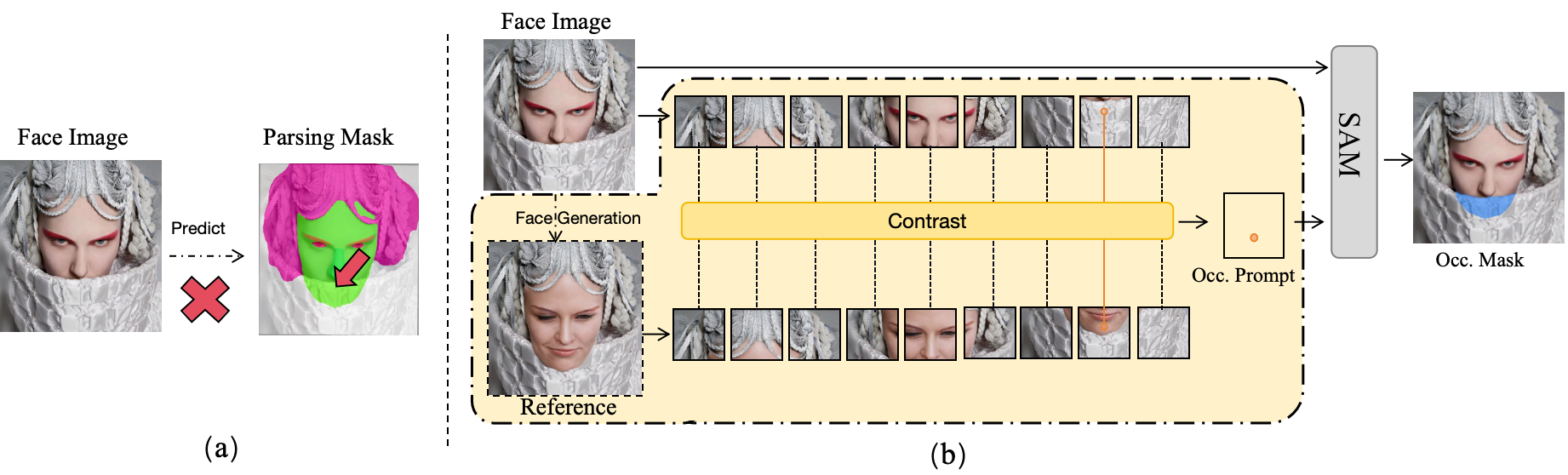}
    \caption{Motivation and Idea Overview. (a) The illustration of the limitations of the current face-parsing algorithms when occlusions occur, which will lead to bad result in downstream tasks. (b) The demonstration of our brief idea and strategy of generating an appropriate prompt for SAM to accurately create an occlusion mask.}
    \label{fig:idea}
\end{figure*}

Face parsing, a task of assigning pixel-level labels to distinguish facial components (\eg, eyes, nose, and mouth), produces critical masks that enable numerous facial manipulation applications including face swapping~\cite{baliah2024realistic,faceswapper}, face editing~\cite{lee2020maskgan}, and facial makeup~\cite{wan2022facial}. However, current face parsing approaches fundamentally misunderstand occlusions by erroneously classifying them as facial regions (Fig.\ref{fig:idea} a). 

The problem lies in the definition of occlusion is considerably high-level. It does not refer to any specific category of object but is a description of spatial relationship between face and other objects. Therefore, it is extremely hard to construct a real-world data covering all possible occlusion object. In addition, to obtain high-quality of occlusion mask is time-consuming, especially in large dataset. Some attempts~\cite{voo2022delving} are conducted to handle these challenges by synthesizing occluded face. However, its generalizability is moderate (Fig.~\ref{fig:front}). 

Recently, visual foundation model, SAM~\cite{kirillov2023segment}, presents a generic segmentation capabilities and a novel interactive paradigm. Due to these merits, it is natural to adopt SAM as a feature representation model for face occlusion segmentation. But how to generate accurate prompts without ground truth mask becomes a thorny issue. Modern face generation model~\cite{richardson2021psp} can produce plausible facial structure and texture even in occluded face, so that occlusion region is apparent by visually contrasting raw and generated images.  

Inspired by this observation, we come up with a totally different solution compared to previous ones. The brief idea, depicted in Fig.~\ref{fig:idea} b, is that given a partially occluded face, a reference face with on occlusion is first produced by a face generation model and spatially compared with the raw face to find different regions. The pixel locations in these regions are used as prompts and coupled with the reference image to be forwarded into SAM to parse out occlusion mask. 

Based on this idea, we propose a concise and effective framework S$^3$POT, Segmentation via Self-Supervised Prompting Occlusion with conTrast, comprising three novel modules: Reference Generation (RG), Feature Enhancement (FE), and Prompt Selection (PS).
RG creates an occlusion-free version of the original face by using a raw face image and its parsed mask as conditions. It prioritizes preserving the facial geometry while authentically restores occluded regions.
To precisely generate occlusion mask, the prompts input to SAM should contain both positives (occlusion) and negatives (face), so as to provide complementary information. However, the feature representation from SAM image encoder is insufficient to locate these prompts by contrasting. Therefore, a two-stage prompt construction strategy is proposed to alleviate this problem. 
At the first stage, FE selects a pixel location of the most similar region as an initial prompt and refines the original and reference image features by a series of prompt-image cross-attention operations.
At the second stage, PS applies a greedy matching algorithm to automatically select a set of positive and negative prompts. A self-attention network, in conjunction with three novel objective functions, are then used to adaptively reweight the prompts based on the underlying characteristics of the prompts.

Our contributions are summarized as following: 
\begin{itemize} 

    \item A concise and effective contrast-driven face occlusion segmentation framework is proposed to deal with infinite occlusion variability.
    \item  A two-stage prompt selection strategy and three novel objective functions are proposed to screening out useful prompts for occlusion segmentation.

    \item We collect a dedicated dataset to evaluate our method and conduct extensive and insightful experiments to demonstrate the superiority of our framework, effect of each component, and underlying principles of method design. 

\end{itemize}

%% file: sec/2_relatedwork.tex
\section{Related Work}
\begin{figure*}
    \centering
    \includegraphics[width=0.99\linewidth]{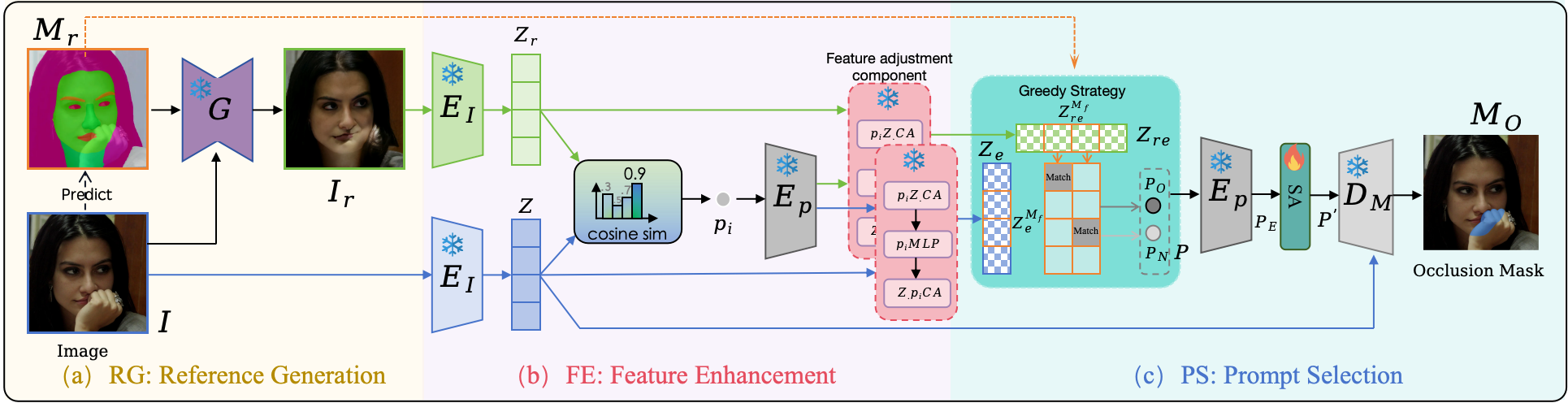}
    \caption{\textbf{The architecture of the proposed framework.} 
    The framework consists of three modules: RG, FE, and PS.
    \textbf{RG:} Given a face image $I$, a corresponding face parsing mask $M_r$ is first predicted and forwarded into a face generator $G$. The original image $I$ is used as a condition in $G$ for producing a reference face $I_r$.
    \textbf{FE:} Both $I_r$ and $I$ are fed into the SAM image encoder $E_I$ to obtain the image tokens $Z_r$ and $Z$, respectively. The cosine similarity between $Z_r$ and $Z$ is used to generate an initial prompt $p_i$. This prompt is first fed into prompt encoder $E_P$ and then paired separately with $Z_r$ and $Z$ before being processed by the feature adjustment component, resulting in enhanced tokens $Z_{re}$ and $Z_{e}$. The feature adjustment component incorporates multilayer perceptron (MLP), prompt-to-image cross-attention ($p_iZ$CA), and image-to-prompt cross-attention ($Zp_i$CA).
    \textbf{PS:} Using $M_r$, the facial region is selected from the enhanced tokens, producing $Z_{re}^{M_f}$ and $Z_{e}^{M_f}$. Then, the cosine similarity between $Z_{re}^{M_f}$ and $Z_{e}$ is computed and a greedy strategy is applied to find similar tokens in $Z_{e}^{M_f}$. Unpaired tokens in $Z_{e}^{M_f}$ are identified as occlusion prompts $P_O$, while the paired tokens constitute the non-occlusion prompts $P_N$. These prompts are processed together by the prompt encoder $E_P$ to obtain $P$. Following a self-attention screening layer (SA), the re-weighted prompt $P'$ is paired with $Z$ and forwarded to the SAM mask decoder $D_M$ to predict the occlusion mask $M_O$.
    }
    \label{fig:method}
\end{figure*}

\noindent
{\bf Face Parsing.}  
Early works in face parsing leverage deep convolutional networks to segment predefined facial components (e.g., eyes, nose, etc.)~\cite{warrell2009labelfaces,khan2015multi,liang2015deep,lin2019face,liu2017face}. Local-based methods focused on predicting each facial part individually by training separate models for different regions. For example, HFP~\cite{luo2012hierarchical} employes a hierarchical approach for parsing individual components, while icNN~\cite{zhou2015interlinked} proposes using multiple CNNs at various scales, fusing them through an interlinking layer to integrate both local and contextual cues. However, due to annotation biases in the training data, these methods inevitably misclassify occlusions as facial regions~\cite{karras2019style,lee2020maskgan}. More recent approaches, such as DML-CSR~\cite{zheng2022decoupled}, introduce a dynamic dual graph convolutional network to address spatial inconsistencies and use cyclic self-regulation to mitigate errors caused by incorrect occlusion labels. In addition, Segface~\cite{narayan2024segface} leverages a lightweight transformer decoder with learnable task-specific tokens to improve segmentation performance for long-tail classes. Voo\cite{voo2022delving} propose to perform occlusion-aware face segmentation by generating synthetic occlusion data. But gap between synthesized data and real-world data is hard to bridge.

\noindent
{\bf SAM-Based Semantic Segmentation.}  
The Segment Anything Model (SAM)~\cite{kirillov2023segment} has demonstrated impressive zero-shot segmentation capabilities using prompts, attributed to its strong feature representation. Several studies have extended SAM for semantic segmentation tasks. OV-SAM~\cite{yuan2024open}, for instance, combines SAM with CLIP~\cite{radford2021learning} to achieve open-vocabulary semantic segmentation, while PerSAM and PerSAM-F~\cite{zhang2023personalize} exploit SAM for personalized segmentation with one-shot guidance. Matcher~\cite{liu2023matcher} utilizes a training-free SAM-based architecture to deliver competitive results. Furthermore, VRP-SAM~\cite{sun2024vrp} introduces an external Visual Reference Prompt Encoder to automatically generate prompts from reference images, and GF-SAM~\cite{zhang2025bridge} offers an efficient, training-free method without the need for external hyperparameters. Despite their successes, these techniques still struggle to detect unknown and diverse occlusions.


\noindent
{\bf Difference.} Different from all previous works, we formulate the face occlusion segmentation task as a contrast-driven prompt generation task. In this way, we can utilize the general feature representation capability of fundation model and the adaptive feature modification characteristic of prompt to achieve self-supervised occlusion segmentation. Thus, it alleviates the effort for data collection and annotation.

%% file: sec/3_method.tex
\section{Method}
\subsection{Reference Generation}
To enable reliable face occlusion segmentation through image contrast, our reference image must strictly preserve geometric consistency with the original image while removing any occlusions. We achieve this using a dual-conditioning approach based on Regional GAN Inversion (RGI) from E4S~\cite{richardson2021psp}, where the face image $I$ provides the texture details and its corresponding parsing mask $M_r$
governs the image structure.
\subsection{Feature Enhancement}
Given an original face $I$ and a reference face $I_r$ as input to the SAM image encoder $E_I$, we obtain the corresponding image tokens $Z$ and $Z_r$. Directly contrasting them cannot yield suitable prompts to indicate occlusion regions, because of the indistinguishability of face and occlusion in current feature space.

To differentiate them in feature space, we exploit the dynamic feature encoding characteristics of prompt and image cross-attention in SAM. Specifically, we first calculate the cosine similarity between \( Z_r \) and \( Z \) and select the pixel location of a token with the highest similarity value as an initial prompt $p_i$. The reason of choosing the most similar location rather than lowest similarity of location is that the reference face is simply same with the original face in major characteristics not all attributes (\eg, illumination), so that locations with lower similarity are prone to have larger probability of containing noise than higher ones (more analysis in ablation study).

The initial prompt \( p_i \) is then passed to a prompt encoder to generate a prompt token. It is combined with \( Z_r \) and \( Z \) separately to be forwarded into a feature adjustment component.  
\textit{Prompt-Image Cross-Attention ($p_iZ$CA)} allows the prompt embedding to attend image features, facilitating the extraction of relevant facial features corresponding to specific coordinate. \textit{Prompt Multilayer Perceptron ($p_i$MLP)} performs nonlinear transformations on each prompt to further enrich their representations. \textit{Image-Prompt Cross-Attention ($Zp_i$CA)} enables image features to attend back to the prompt embeddings, enriching the contextual information.

The inspiration of the structure and the weights of the component are both from the SAM mask decoder. Thus, benefiting from the robust and pretrained feature fusion capabilities, the output of FE are enhanced image tokens \( Z_{re} \) and \( Z_{e} \). This bidirectional fusion results in more expressive and discriminative features, leading to more sharp contrast between facial regions and occlusions.
\subsection{Prompt Selection}
\begin{algorithm}[t]
\caption{Greedy Matching Algorithm}
\label{alg:optimized_greedy}
\begin{algorithmic}[1]
\REQUIRE Image features \footnotesize{$Z_{e} \in \mathbb{R}^{hw \times d}$}, Reference subset \footnotesize{$Z_{re}^{M_f} \in \mathbb{R}^{N({Z_{re}^{M_f}})\times d}$}, Image subset \footnotesize{$Z_{e}^{M_f} \in \mathbb{R}^{N({Z_{e}^{M_f}})\times d}$}
\ENSURE Non-occ. prompts $P_N$, Occ. prompts $P_O$
\STATE \textbf{Vectorized Similarity Computation:}
\STATE Sim-Matrix $S = \hat{Z}_{e} (\hat{Z}_{re}^{M_f})^{T} \in \mathbb{R}^{N(Z_{e}) \times N({Z_{re}^{M_f}})}$
\STATE \textbf{Initialize Masks:}
\STATE $rows \leftarrow [\text{True}]_{N({Z_{e}})}$, $cols \leftarrow [\text{True}]_{N({Z_{re}^{M_f}})}$
\STATE \textbf{Greedy Matching:}
\STATE $matched\_pairs \leftarrow []$
\FOR{$k = 1$ to $N({Z_{re}^{M_f})}$}
    \STATE Find $(i^*, j^*) = argmax_{i,j} S[i,j]$ where $rows[i] \land cols[j]$
    \STATE Append $(i^*, j^*)$ to $matched\_pairs$
    \STATE Set $rows[i^*] \leftarrow \text{False}$, $cols[j^*] \leftarrow \text{False}$
\ENDFOR
\STATE \textbf{Identify Feature-space Prompts:}
\STATE $P_N \leftarrow \{i \mid (i,j) \in matched\_pairs,\, i \in M_f\}$ 
\STATE $P_O \leftarrow \{0, 1, ..., N(Z_e^{M_f})-1\}\setminus P_N$ 
\STATE \textbf{Return} $P_N$, $P_O$ (as feature map indices)
\end{algorithmic}
\label{alg:greedy}
\end{algorithm}

Providing only non-occlusion prompt or occlusion prompt can lead to inaccurate segmentation granutlarity, as mask decoder lacks information about complementary samples. Therefore, the purpose of this module is to construct a set of informative positive prompts (occlusion) and negative prompts (face) for mask decoder. Considering the diversity and unknown nature of occlusion, we propose a global-local-global (GLG) strategy. 

Focusing on the facial region firstly, we define the facial region set as $F = \big\{ \text{lip}, \text{eyebrows}, \text{eyes}, \text{nose}, \text{skin}, \text{mouth} \big\}.$
We construct the facial region mask by selecting regions through label matching:
$M_f = \big\{ m_i \mid m_i\in M_r \cap F \big\}.$
Subsequently, we extract the subset of image tokens $\mathbf{Z}_{\mathrm{re}}^{M_f}$ where the facial region mask $M_f$ is active:
$\mathbf{Z}_{\mathrm{re}}^{M_f} = \big\{ \mathbf{z}_i \mid \mathbf{z}_i \in \mathbf{Z}_{\mathrm{re}} , m_i = 1 \big\},$
This ensures the subsequent comparison process concentrates on relevant facial features.
We then use $Z_{e}$, $Z_{re}^{M_f}$ and $Z_{e}^{M_f}$ do the greedy matching algorithm to select the most similar image token pairs, resulting in occlusion prompt $P_O$ and non-occlusion prompts $P_N$. The algorithm execution flow can be seen in the pseudocode Alg.~\ref{alg:greedy}.
We combine \( P_N \) and \( P_O \) into a new prompt set $P$ and feed them into prompt encoder to get the embedding $P_E = E_P(P)$.



\begin{figure}
    \centering
    \includegraphics[width=0.6\linewidth]{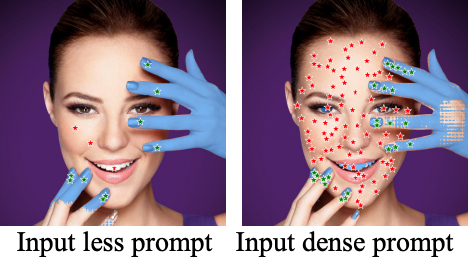}
    \caption{Motivation for filtering redundancy and noise in the new prompt set $P$ . Even with a large number of accurate prompts input to SAM, performance does not improve. A smaller set of well-chosen, accurate prompts is sufficient for optimal results.}
    \label{fig:sa_motivation}
\end{figure}

Despite the enhanced feature, \( P \) may still contain noise and redundant information. As shown in Fig.~\ref{fig:sa_motivation}, providing an excessive number of prompts to SAM does not yield better performance and can even degrade results. To address this, we introduce a self-attention mechanism to automatically filter the prompts, ensuring that only the necessary information is retained. This is accomplished by computing correlation between each element and dynamically assigning it to each prompt.   
The re-weighted prompts \( P' \) are then fed into the SAM mask decoder along with the image features \(Z\) to generate the predicted mask \( M_O \).
\begin{figure}[t] 
    \centering
    \includegraphics[width=0.6\linewidth]{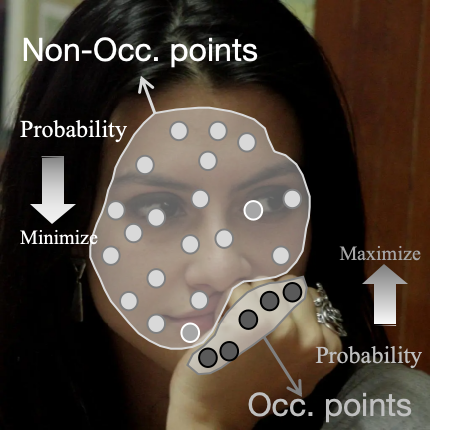}
    \caption{The Illustration of the intuition of objective functions.
        The prompt points are categorized into two groups: occlusion points (black) and non-occlusion points (grey). In the predicted mask probability map, the probability values of occlusion points should be heightened, while those of non-occlusion points should be suppressed. When reducing the probabilities of non-occlusion points, in addition to lowering the average probability, we also introduce a penalty term to prevent individual points from exhibiting anomalously high values.}   
    \label{fig:loss}
\end{figure}

\noindent\textbf{Objective Function Design.} 
In the absence of ground truth of occlusion region, we have developed three novel objective functions (Occlusion Prompt Recall (OPR), Face Prompt Recall (FPR), Face Prompt Penalty (FPP)) to regulate the self-attention learning by exploiting underlying characteristics of prompt set $P$.  

\noindent
\textit{OPR.} As shown in Fig.~\ref{fig:loss}, occlusion prompt (\ie, black points) should have higher value in the predicted probability map from SAM mask decoder. This characteristic is captured by the occlusion prompt recall function
\begin{equation}
    \mathcal{L}_{\text{recall\_occ}} = -\frac{1}{N_o} \sum_{j=1}^{N_o} \log q_j,
    \label{eq:loss_recall_occ}
\end{equation}
where \( N_o \) is the number of occlusion prompt, and \( q_j \) is the predicted probability of the \( j \)-th prompt. However, relying solely on this function can cause the mask to become excessively large, as it does not ensure the exclusion of non-occluded areas (\ie, face regions). Therefore, it is necessary to introduce additional constraints for the non-occluded regions.

\noindent
\textit{FPR.} Non-occlusion prompt (grey points Fig.~\ref{fig:loss}) should have lower probability value. To enforce this, we define a face prompt recall function as:
\begin{equation}
    \mathcal{L}_{\text{recall\_face}} = \frac{1}{N_f} \sum_{i=1}^{N_f} \log q_i,
    \label{eq:loss_recall_face}
\end{equation}
where \( N_f \) denotes the number of face prompt, and \( q_i \) represents the predicted value of the \( i \)-th face prompt in the probability map. It servers as a complementary term for $\mathcal{L}_{\text{recall\_occ}}$ to constrain the predicted mask to expand to non-occlusion region in certain degree.

\noindent
\textit{FPP.} While \(\mathcal{L}_{\text{recall\_face}}\) aims to lower the average probability of face prompt, it may not sufficiently penalize points with high probability. To address this, 
we introduce a face penalty function:
\begin{equation}
    \mathcal{L}_{\text{face\_penalty}} = \frac{1}{N_f} \sum_{i=1}^{N_f} \sigma\left(\alpha \left(q_i - 0.5\right)\right),
    \label{eq:loss_face_penalty}
\end{equation}
where \( \sigma \) denotes the sigmoid function, \( \alpha \) is a scaling factor that controls the steepness of the sigmoid function, and \( q_i \) is the predicted probability of the \( i \)-th face point. This term pushes \( q_i \) to be less than 0.5, especially the points with higher probability. Thus, it can mitigate the risk of misclassifying face regions as occlusion (the bright grey point at lip in Fig.~\ref{fig:loss}).

The self-attention network is effectively trained by minimizing the overall function of the aforementioned ones, :
\begin{equation}
    \mathcal{L}_{\text{total}} = \mathcal{L}_{\text{recall\_face}} + \mathcal{L}_{\text{recall\_occ}} + \lambda \mathcal{L}_{\text{face\_penalty}},
    \label{eq:loss_total}
\end{equation}
where \( \lambda \) is a hyperparameter that balances the contribution of the face penalty relative to the recall terms.

%% file: sec/4_experiment.tex
\section{Experiments}

\begin{figure*}
    \centering
    \includegraphics[width=1\linewidth]{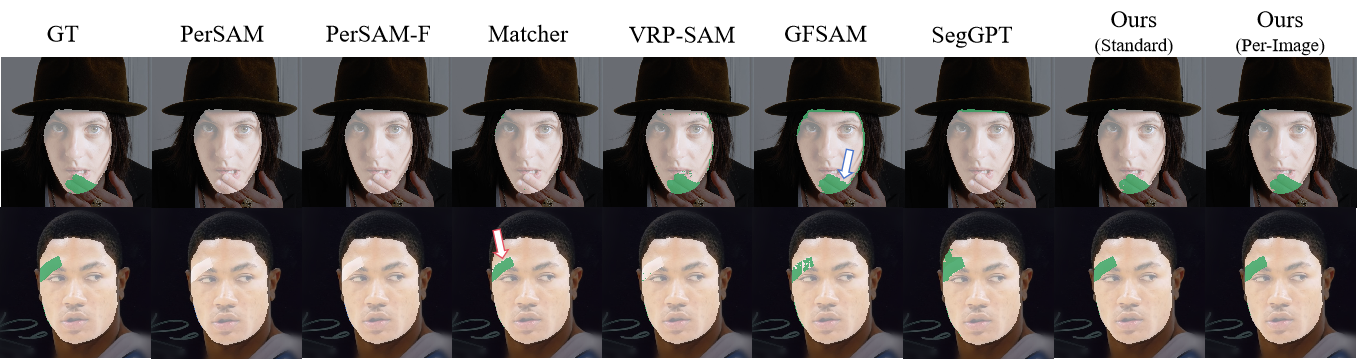}
    \caption{Visual comparison with visual reference segmentation models. White indicates the complete mask and green denote occlusions.}
    \label{fig:comparsion}
\end{figure*}


\begin{figure}
    \centering
    \includegraphics[width=0.9\linewidth]{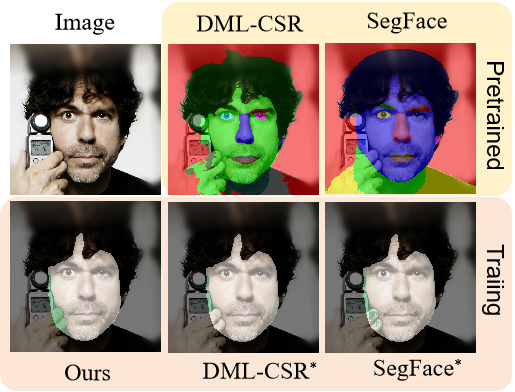}
    \caption{Visual comparison with face parsing models. For pretrained models, we display the parsing results. For models trained on our data, we present the segmentation results, with the green region indicating the predicted occlusion area.}
    \label{fig:parsing}
\end{figure}

\begin{figure}
    \centering
    \includegraphics[width=1\linewidth]{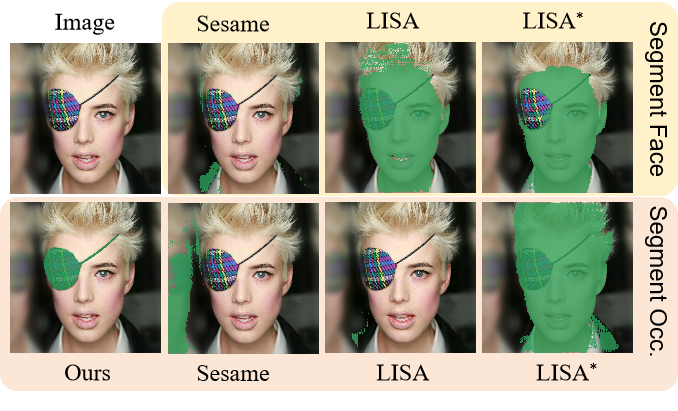}
    \caption{Visual comparison with text-guided reasoning segmentation models. These models are capable of segmenting faces or occlusions based on different prompts. Images with a yellow background represent face segmentation results, while those with a orange background correspond to occlusion segmentation. The green regions indicate the predicted masks for the segmentation target.}
    \label{fig:LLMcomparison}
\end{figure}

\begin{figure*}[t]
    \centering
    \includegraphics[width=0.95\linewidth]{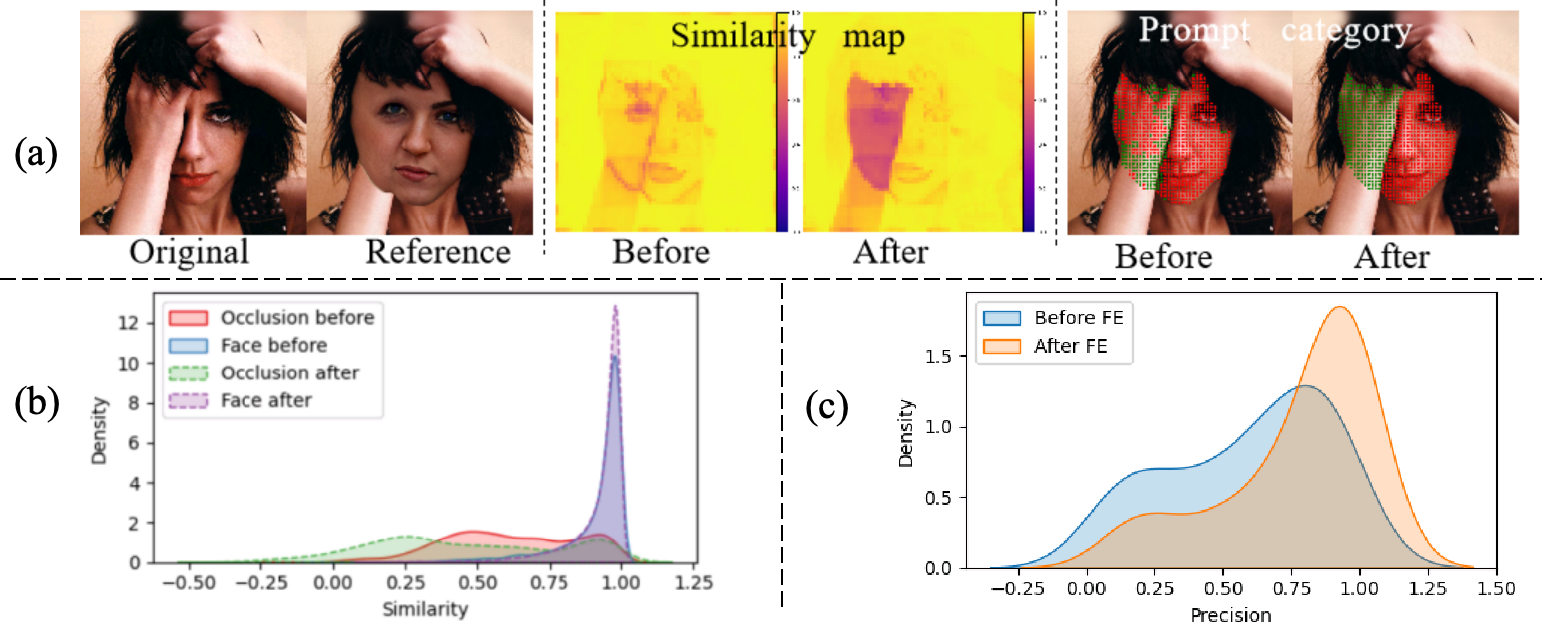}
    \vspace{-0.2cm}
    \caption{Illustration of the effect Feature Enhancement. (a) Similarity map is obtained by pixel-wise computing cosine distance between feature maps of original and reference images. Prompt category is occlusion prompt (green points) and non-occlusion prompt (red points), generated by matching feature maps of reference and original images using greedy strategy. Matched location is treated as occlusion prompt. The Unmatched is non-occlusion prompt. (b) Kernel density estimation (KDE) of similarity between face and occluded regions before and after FE. (c) KDE of precision distribution of the occlusion prompt before and after applying FE.}
    \label{fig:enhance}
\end{figure*}

\noindent
\textbf{Datasets.}
To construct a face occlusion dataset for algorithm evaluation, we leverage the capabilities of large language models to screen out occluded face images from two face datasets CelebAMask-HQ~\cite{lee2020maskgan} and FFHQ~\cite{karras2019style}. Specifically, we utilize Qwen~\cite{bai2023qwen} with carefully designed language prompts to filter and select faces containing occlusions. Subsequently, due to the lack of precise masks, we employ the annotation tool X-AnyLabeling~\cite{X-AnyLabeling} to generate accurate masks for each selected image. This process results in 1,389 images from CelebA and 1,104 images from FFHQ. Out of these, 104 images from FFHQ and 96 images from CelebA are designated for validation. The remaining FFHQ images are allocated for training, while the rest of the CelebA images are reserved for testing. This dataset will be released in the future.

\noindent
\textbf{Implementation Details.}  
Regional GAN Inversion (RGI) technology from E4S~\cite{liu2023fine} for mask-guided face generation is adopted in reference generation. The feature adjustment component is initialized with the weights from the counterpart of SAM Mask Decoder. In our objective functions, the hyperparameter $\lambda$ is set to 0.5, and the scaling factor $\alpha$ is set to 10.
Our framework is implemented in PyTorch and trained on 6 NVIDIA RTX6000 GPUs for 50 epochs with an initial learning rate of $1 \times 10^{-4}$ and a batch size of 3. AdamW optimizer~\cite{loshchilov2017decoupled} is used with a cosine learning rate decay schedule. All input images are resized to $1,024 \times 1,024$ pixels. 
We have two experiment settings. 
\textit{Standard Training:} A single model is trained on training data and evaluated on test data.
\textit{Per-Image Overfitting:} A separate model is trained for each image in test data.
S$^3$POT is exclusively compared to previous works using  intersection over union (IoU) of occlusion region (Occ. IoU).

\subsection{Comparison with State-of-the-art Methods}
\begin{table}[t]
\centering
\resizebox{0.47\textwidth}{!}{%
\begin{tabular}{lccccc}
\hline
Method       &Venue         & Seg. Target & TR & SPV & Occ. IoU $\uparrow$\\ \hline
\multicolumn{5}{l}{\textit{face parsing}}                    \\
SegFace\textsuperscript{*}~\cite{narayan2024segface}&\gray{ AAAI25}   & face   &$\checkmark$ &$\checkmark$ &73.1 \\
SegFace~\cite{narayan2024segface}&\gray{ AAAI25}   & face   &$\times$ &$\checkmark$ & 58.2\\
DML-CSR\textsuperscript{*}~\cite{zheng2022decoupled}&\gray{ CVPR22}   & face   & $\checkmark$& $\checkmark$ & 72.4\\
DML-CSR~\cite{zheng2022decoupled}&\gray{ CVPR22}   & face   & $\times$& $\checkmark$ &70.6 \\ \hline
\multicolumn{5}{l}{\textit{visual reference segmntation}}    \\
SegGPT~\cite{wang2023seggpt}&\gray{ ICCV23} & face           & $\times$&$\checkmark$  &47.1\\
PerSAM~\cite{zhang2023personalize}&\gray{ ICLR24}                & face   &$\times$&$\times$  & 21.3  \\
PerSAM-F~\cite{zhang2023personalize}&\gray{ ICLR24}  & face          &$\checkmark$ & $\checkmark$ &  22.4 \\
Matcher~\cite{liu2023matcher}&\gray{ ICLR24} & face           &$\times$ &$\times$ & 46.6 \\
VRP-SAM~\cite{sun2024vrp}&\gray{ CVPR24} & face          &$\checkmark$ & $\checkmark$ & 47.4 \\
GFSAM ~\cite{zhang2025bridge}&\gray{ NeurIPS24} & face           &$\times$  &$\times$ &51.2   \\ \hline
\multicolumn{5}{l}{\textit{text-guided reasoning segmentation}}        \\
Sesame~\cite{wu2024see}&\gray{ CVPR24}  & occlusion      &$\times$& $\checkmark$ &  6.72 \\
Sesame~\cite{wu2024see}&\gray{ CVPR24} & face         &$\times$  & $\checkmark$ &17.2\\ 
LISA~\cite{lai2024lisa}&\gray{ CVPR24}     & occlusion   &$\times$   &$\checkmark$ &6.47   \\
LISA~\cite{lai2024lisa}&\gray{ CVPR24}     & face         &$\times$  & $\checkmark$&  35.2\\
LISA\textsuperscript{*}~\cite{lai2024lisa}&\gray{ CVPR24}     & occlusion   &$\checkmark$   &$\checkmark$ &10.8  \\
LISA\textsuperscript{*}~\cite{lai2024lisa}&\gray{ CVPR24}     & face         &$\checkmark$  &$\checkmark$ & 43.1 \\\hline
\multicolumn{5}{l}{\textit{contrast-based segmentation}}             \\
Ours (Standard)    &-    & occlusion   &$\checkmark$   & $\times$ &  {\ul76.2} \\
Ours (Per-Image)   &-   & occlusion    &$\checkmark$  & $\times$ &  \textbf{80.4}\\ \hline
\end{tabular}
}
\caption{Segmentation results of comparing our method with previous works. The methods fall into three categories: face parsing, visual reference segmentation, and text-guided reasoning segmentation. While most existing methods only segment the face, we use a complete face mask to transform the face segmentation into an occlusion mask. SPV indicates whether this method falls under supervised learning, while TR denotes whether we train this model in our dataset. The trained methods are marked with *.}
\label{tab:comparison}
\end{table}

Occlusion segmentation has received limited attention in the literature, resulting in few directly comparable prior works. Nevertheless, several related tasks—although not originally developed for occlusion segmentation—can be adjusted to address occlusion. We compare our method with state-of-the-art works in related fields, including face parsing, visual reference segmentation, and reasoning segmentation. For a fair evaluation, the output from face segmentation methods is converted to occlusion mask using full face mask. All comparisons are conducted on the test data.

\noindent
\textbf{Comparison with Face Parsing Models.} Face parsing models focus on segmenting predefined facial components, typically excluding occlusion attributes and only addressing elements such as hats, clothing, and background. In this work, we compare our approach against two models: DML-CSR~\cite{zheng2022decoupled} and SegFace~\cite{narayan2024segface}. We train them on our dataset with annotated grounnd-truth, but modify them to perform full-face segmentation instead of multi-label prediction, resulting in SegFace\textsuperscript{*} and DML-CSR\textsuperscript{*}. We also use their pretrained models for comparison.
Results are presented in Table \ref{tab:comparison}. SegFace\textsuperscript{*} achieves the IoU of 73.1, while DML-CSR\textsuperscript{*} reaches an IoU of 72.4. Both of them are much lower than our results. A visual comparison is provided in Fig.~\ref{fig:parsing}. Notably, our method fully segments the both object and hand. Pretrained models cannot solve occlusion problems. DML-CSR\textsuperscript{*} only captures the boundary of occlusion, and SegFace\textsuperscript{*} still ignore some part of fingers.

\noindent
\textbf{Comparison with Visual Reference Segmentation.} Most approaches in this domain use one-shot segmentation, meaning that a support image-mask pair is required during the inference stage. We compare our method with several state-of-the-art models, including SegGPT~\cite{wang2023seggpt}, PerSAM~\cite{zhang2023personalize}, GFSAM~\cite{zhang2025bridge}, Matcher~\cite{liu2023matcher}, PerSAM-F~\cite{zhang2023personalize}, and VRP-SAM~\cite{sun2024vrp}. Since occlusions cannot be explicitly defined as a distinct object type, we employ them to segment only the unoccluded areas of the face to assess whether they can correctly exclude occlusions from face segmentation.
PerSAM~\cite{zhang2023personalize}, Matcher~\cite{liu2023matcher}, and GFSAM~\cite{zhang2025bridge} are training-free methods so that they can be directly adopted to output occlusion mask.
PerSAM-F~\cite{zhang2023personalize} is fine-tuned on a specific object class to identify similar objects. In our experiments, we provide it with an occluded face and the corresponding ground truth face mask for fine-tuning. The learned weights are then used to perform inference on test data. VRP-SAM~\cite{sun2024vrp} needs to be trained on our training data before evaluated on test data. Since SegGPT does not provide training code, we use its pretrained model for comparison. The results in Tab.~\ref{tab:comparison} show that our method achieves the best results.

As depicted in Fig.~\ref{fig:comparsion}, PerSAM, PerSAM-F, and SegGPT tend to merge the finger area with portions of the face near the lips due to their similar skin tones. Notably, SegGPT produces a slightly smaller face segmentation along the boundaries, so when converting using a full mask, the region where the face connects to the hat is classified as face occlusion. Similarly, although GFSAM encounters the same issue, it can still segment the fingers. But the finger boundaries are not well defined (as shown by the blue arrow). In contrast, VRP-SAM generally segments the overall finger area well; however, it sometimes produces small, scattered regions and stray points. Under our standard experimental mode, a small gap appears in the mask, but the PerImage experimental mode effectively resolves this problem and yields the best results.

When segmenting a band-aid, PerSAM, PerSAM-F, and VRP-SAM fail to distinguish it. GFSAM detects only a small portion of the band-aid, while SegGPT mistakenly classifies the adjacent skin as an occlusion. Aside from our method, Matcher achieves the best results, although its boundaries are relatively rough (as shown by the red arrow).

\noindent
\textbf{Comparison with Reasoning Segmentation Models.}  
Reasoning segmentation requires not only extracting localized cues but also employing complex reasoning to interpret implicit user intentions without relying on predefined categories. 
In our study, we compare the proposed approach with two state-of-the-art reasoning segmentation models, LISA~\cite{lai2024lisa} and Sesame~\cite{wu2024see}. Both models are evaluated using two prompts—“Please help me segment unoccluded face” and “Please help me segment face occlusion”—which lead them to pursue different segmentation objectives. Notably, they are the only ones that can handle occlusion.
For a more rigorous comparison, we train LISA on our training dataset and evaluate it on our test dataset, resulting in a variant referred to as LISA\textsuperscript{*}. Additionally, we include results from the pretrained versions of these models to leverage the robust generalization capabilities of large language models.
The quantitative results, presented in Table~\ref{tab:comparison}, indicate that when segmenting occlusions, both LISA and Sesame achieve an IoU of approximate 6.5, suggesting that they do not fully comprehend the concept of occlusion within a linguistic context. In contrast, in segmenting faces setting, LISA’s IoU increases to 35.2 and Sesame’s to 17.2. 
Fig.~\ref{fig:LLMcomparison} shows the qualitative results. Both Sesame and LISA struggle to differentiate between occlusions and facial features. Although LISA\textsuperscript{*} exhibits a clearer definition of the face and achieves better face segmentation performance compared to occlusion segmentation (as it is generally easier to delineate a face), the models still fail to identify certain details, such as the bandage on the eye mask.
\begin{table}[t]
\centering
\begin{tabular}{ccccc|c}
\hline
\multicolumn{1}{c}{\multirow{2}{*}{FE}} &
  \multicolumn{4}{c|}{PS} &
  \multicolumn{1}{c}{\multirow{2}{*}{IoU$\uparrow$}} \\ 
 \multicolumn{1}{c}{} &
  \multicolumn{1}{c}{TH} &
  \multicolumn{1}{c}{Otsu.} &
  \multicolumn{1}{c}{GS} &
  \multicolumn{1}{c|}{SA} &
  \multicolumn{1}{c}{} \\ \hline
   -& -& -& \checkmark& -& 51.2\\
   -& -& -& \makecell{\checkmark} & \makecell{\checkmark} & {\ul 68.3}\\
\makecell{\checkmark} & \makecell{\checkmark} &- & -& -& 17.9\\
\makecell{\checkmark} & -& \makecell{\checkmark} & -& -& 25.6\\
\makecell{\checkmark} & -& -& \makecell{\checkmark} & -& 56.9\\
\makecell{\checkmark} & \makecell{\checkmark} & -& -& \makecell{\checkmark} & 34.5\\
\makecell{\checkmark} & -& \makecell{\checkmark} & -& \makecell{\checkmark} & 38.1\\ 
\makecell{\checkmark} & -& -& \makecell{\checkmark} & \makecell{\checkmark} & \textbf{76.2}\\ \hline
\end{tabular}
\centering
\caption{ Ablation study results of feature enhancement (FE) and prompt selection (PS). TH denotes threshold. GS denotes greedy strategy.}
\label{tab:ablation}
\end{table}

\begin{table}[t]
\centering
\begin{tabular}{ccc|c}
\hline
$\mathcal{L}_{\text{recall\_face}}$&$\mathcal{L}_{\text{recall\_occ}}$& $\mathcal{L}_{\text{face\_penalty}}$& IoU$\uparrow$     \\ \hline
\makecell{\checkmark}  &  -&   -&  7.54\\
-&  \makecell{\checkmark} &    -&  16.6\\
-&  -& \makecell{\checkmark} &  7.32\\
\makecell{\checkmark} & \makecell{\checkmark} & -&  {\ul 73.5} \\
\makecell{\checkmark} &  -&  \makecell{\checkmark} &  7.29\\
 -&  \makecell{\checkmark} & \makecell{\checkmark} & 72.3 \\ 
\makecell{\checkmark}& \makecell{\checkmark} & \makecell{\checkmark} &  \textbf{76.2} \\ \hline    

\end{tabular}
\caption{ Ablation study on different objective functions. }
\label{tab:loss}
\end{table}
\subsection{Ablation Study}
\noindent
\textbf{Effect of Feature Enhancement.}
As presented in Tab.~\ref{tab:ablation}, the integration of the Feature Enhancement (FE) leads to a substantial improvement in IoU (68.3 to 76.3). 
To further demonstrate the improvement is caused by FE, we computed the cosine similarity between the features before and after applying FE as shown in Fig.~\ref{fig:enhance}(a). 
The similarity within the hand regions significantly decreases after applying FE. The location of occlusion prompts (red points) and non-occlusion prompts (green points) changes dramatically after FE. These phenomena indicate an enhanced distinction between occlusion and face in feature space.

In Fig.~\ref{fig:enhance}(b), we analyze the similarity distribution of image features before and after FE. As shown, the enhanced features emphasize more face-related representation, resulting in the similarity scores within the face region being more tightly clustered around 1. Additionally, this enhancement reduces the similarity of occlusion regions. Specifically, while the original occlusion regions exhibited similarity scores clustered around 0.5, the enhanced features shift this cluster center to approximately 0.25. Consequently, the enhanced features demonstrate an improved ability to discriminate occlusions and faces.
To demonstrate the effect of FE on occlusion prompt generation, we compute the precision distribution of the prompt predicted by using greedy strategy on features with or without FE. The distribution is presented in Fig.~\ref{fig:enhance}(c). The results indicate a significant improvement in correctness of occlusion prompt when enhanced feature is employed, with the probability distribution of precision scores shifting closer to from 0.75 to 1. This can explain why our designed objective functions can provide supervision information. 
The underlying reason is the initial prompt $p_i$ (face prompt) interacts with image feature by a series of cross-attention operation in feature ajustment component to modify the feature so that it is prone to discriminate face and non-face region.

\noindent
\textbf{Effect of Prompt Selection.}
Directly removing this component and using $p_i$ as prompt to obtain occlusion mask is inappropriate, as $p_i$ is face prompt not occlusion one. To assess the necessity of this component and current implementation, we conduct experiment with another two prompt selection strategies: a fixed similarity threshold of 0.5 and an adaptive threshold determined using Otsu's method~\cite{otsu}. As shown in Tab.~\ref{tab:ablation}, using a fixed threshold (TH) results in a significant decrease in performance from 56.9 to 17.9 and 76.2 to 34.5, proving it to be non-universal. Additionally, the similarity distribution does not consistently exhibit a bimodal pattern assumed by Otsu's method, resulting in a performance drop from 56.9 to 25.6 and 76.3 to 38.1.
It is notable that after adding SA all results are increased by a large margin. This demonstrates the importance and effectiveness of SA and designed object function. The reason of the improvement is that SA can effectively shrink the redundancy or noise in prompts by adaptively adjust the weight of each prompt. 

\noindent
\textbf{Effect of Objective Functions.}
Utilizing any single term of the overall objective function fails to produce satisfactory results, as $\mathcal{L}_{\text{recall\_occ}}$ and  $\mathcal{L}_{\text{recall\_face}}$ only control the recall for either occluded or non-occluded areas and $\mathcal{L}_{\text{face\_penalty}}$ controls the false positive of predicted occlusion mask. Combining $\mathcal{L}_{\text{recall\_face}}$ and $\mathcal{L}_{\text{face\_penalty}}$ results in 7.29 IoU. Because these two terms lead to occlusion region is under-segmentation. Since $\mathcal{L}_{\text{recall\_occ}}$ and $\mathcal{L}_{\text{recall\_face}}$ encourage predicted occlusion region and non-occlusion region to cover as many occlusion prompts and face prompts respectively as possible, they can works in a complementary way to obtain 73.5 IoU. By adding penalty to face prompts with high prediction value, IoU increases to 76.2.



\begin{figure}
    \centering
    \includegraphics[width=0.8\linewidth]{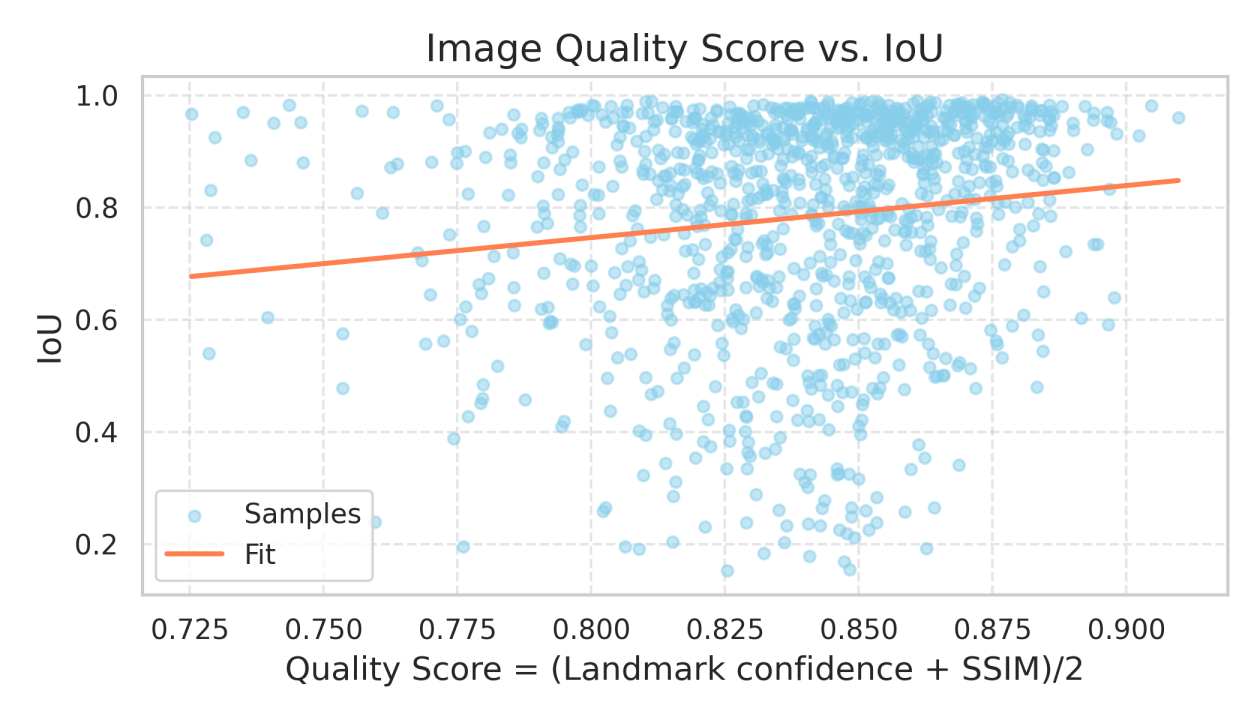}
    \caption{Relation between IoU
and reference image quality.}
    \label{fig:sensitivity}
\end{figure}
\subsection{Analysis on the Quality of the Reference Image}
Our method leverages a generated reference image to guide the restoration of occluded facial regions. A natural question arises regarding the sensitivity of our final performance to the quality of this reference image. To investigate this, we first establish a quantitative metric to evaluate the quality of the generated reference.

We argue that an effective reference image does not require pixel-perfect reconstruction. Instead, it should satisfy two primary criteria: 1) the reconstructed occluded region should be structurally plausible as a facial part, and 2) the non-occluded region should maintain high structural similarity to the original input. Based on these criteria, we propose a composite Quality Score for the reference image. 
\begin{itemize}
    \item To measure the plausibility of the reconstructed occluded region, we use the averaged face landmark detection confidence within that area. A higher confidence score indicates that the generated region is more face-like.
    \item To measure the fidelity of the non-occluded region, we compute the Structural Similarity Index (SSIM) between the generated reference and the original image in that region.
\end{itemize}
The final Quality Score is defined as the mean of these two metrics.
In our experiments, the generated reference images achieve an average SSIM of 0.87 for the non-occluded regions. More importantly, the averaged landmark detection confidence in the occluded regions significantly increases from 0.59 (in the raw occluded image) to 0.81 (in our generated reference). This yields a representative Quality Score of $(0.87 + 0.81) / 2 = 0.84$.
Crucially, we experimentally demonstrate that our method's final output is robust to variations in this Quality Score. As qualitatively shown in Fig.~\ref{fig:sensitivity}, our method produces promising segmentation results even when the reference image's quality is not optimal. This analysis confirms that our approach effectively utilizes the reference for high-level structural guidance rather than depending on its pixel-level perfection, highlighting the robustness of our proposed mechanism.

%% file: sec/5_conclusion.tex
\section{Conclusion}
We propose S$^3$POT, a contrast-driven and self-supervised occlusion segmentation framework for precise mask prediction. By developing occlusion-free reference faces via structural-consistent generation, enhancing discriminative features through attention-based reweighting, and designing three novel spatial-semantic objective functions, our method resolves occlusion ambiguity without requiring labeled training data. Comprehensive experiments demonstrate S$^3$POT outperforms existing face parsing models while effectively preserving facial integrity in occluded regions. The framework's modular design and foundation model compatibility enable seamless integration with existing facial analysis pipelines. While our method effectively handles common occlusions, extremely fine grain occlusions (e.g., facial hair) or ambiguous boundaries (e.g., translucent objects) may lead to partial missegmentation. Future work could explore multi-scale feature fusion or uncertainty-aware masking to address these edge cases.